\documentclass{article}
\usepackage{PRIMEarxiv}
\usepackage[utf8]{inputenc} 
\usepackage[T1]{fontenc}    
\usepackage{hyperref}       
\usepackage{url}            
\usepackage{booktabs}       
\usepackage{amsfonts}       
\usepackage{nicefrac}       
\usepackage{microtype}      
\usepackage{lipsum}
\usepackage{graphicx}       
\usepackage{amsmath,amssymb}
\title{Measuring Diversity in Co-creative Image Generation}

\author{
  Francisco Ibarrola\\
  The University of Sydney,\\
  NSW, Australia.\\
  \scalebox{.8}[1.0]{\texttt{francisco.ibarrola@sydney.edu.au}}\\
  \And
  Kazjon Grace \\
  The University of Sydney,\\
  NSW, Australia.\\
  \scalebox{.8}[1.0]{\texttt{kazjon.grace@sydney.edu.au}} \\
}

\begin{document} 
\maketitle  
\begin{abstract}
Quality and diversity have been proposed as reasonable heuristics for assessing content generated by co-creative systems, but to date there has been little agreement around what constitutes the latter or how to measure it. Proposed approaches for assessing generative models in terms of diversity have limitations in that they compare the model's outputs to a ground truth that in the era of large pre-trained generative models might not be available, or entail an impractical number of computations. We propose an alternative based on entropy of neural network encodings for comparing diversity between sets of images that does not require ground-truth knowledge and is easy to compute. We also compare two pre-trained networks and show how the choice relates to the notion of diversity that we want to evaluate. We conclude with a discussion of the potential applications of these measures for ideation in interactive systems, model evaluation, and more broadly within computational creativity.
\end{abstract}


\section{Introduction}


The quality of generative AI text-to-image systems is improving rapidly, whether you assess that by fit-to-prompt, perceived realism, Frech\'et Inception Distance (FID) \cite{heusel2017FID}, or by virtually any other measure. In a computational creativity context, however: is  ``quality'' all we need? Especially in an interactive and/or co-creative context, that seems to be a dangerous assumption, given the long history in our field of creativity definitions featuring at least a duopoly of constituent factors: one broad cluster of factors that have been variously referred to as value, utility, appropriateness or quality, and another broad cluster usually referred to as novelty, originality, or surprise.  The focus on quality is natural given the rapid advances of these technologies, but several specific questions arise when considering the sufficiency of image quality in our context: What happens when the prompt is ill-formed, because a user doesn't yet know what they want?  What happens when a generator is asked to produce something that deviates substantially from the training data? What happens when a generator's understanding of a word or phrase differs from the user's? 

Algorithmic measures of quality do not -- at least at present -- offer a way to address any of those questions. Conversely, using human subjects evaluation is too slow and too expensive to be a feasible source of feedback at the scale required to improve the underlying models. In this paper we propose that, at least for interactive contexts, generator quality (usually defined as some combination of matching the distribution of the data and accurately reflecting any conditioning stimuli such as prompts) must be accompanied by generator diversity (which we broadly define as maximising the breadth of options among the outputs, although we provide a more specific entropy-based definition below).  The importance of generators offering users diverse options has been raised in computational creativity before, particularly in the field of procedural content for games \cite{smith2010analyzing}, we seek to expand those notions to cover all interactive generative AI, at least where output creativity is a potential goal.

Diversity has been proposed \cite{preuss2014searching,ibarrola2023affect} as desirable in co-creative systems in the past, and in this paper we expand on those proposals, arguing for the criticality of diversity measures in interactive generative systems. We contend that the within-set diversity of generated content is a useful counterpart to quality in evaluating interactive text-to-image systems, formalise the problem of measuring it, and then present generalisable algorithms for doing so. By ``within-set diversity'' we refer abstractly to the breadth of a set of responses provided to a user as part of a single ``round'' of generation.  Our motivation for this definition is that 1) creative tasks are by definition ill-specified and effective (human) approaches to combating that typically involve reframing/reformulation \cite{schon1992designing,dorst2015frame,grace2016surprise}, a general finding that has been replicated specifically in the text-to-image literature in the form of iterative prompt ``engineering'' \cite{oppenlaender2022creativity}, 2) a creative system is unlikely to know in advance the direction in which its user might want to reformulate the ``problem'',  and 3) in interfaces where users are presented with multiple options to choose from (common in text-to-image UIs), traditional single-artefact models of novelty or surprise may result in duplicate content. 

Our focus on diversity as a desirable quality for results produced during an on-going creative process is further motivated by research from the Information Retrieval (IR) community, which has long explored the utility of diversity as an accompaniment to accuracy/similarity in retrieving sets of search or recommendation system results \cite{candillier2011diversity,kunaver2017diversity}. In the IR community, the goal is to maximise the chances that an answer to the user's query exists within the top N results -- for a concrete example, consider that ``top N'' to be on the ``first page'' of a search engine. To do so, it's not sufficient to present a set of most-similar or most-accurate results, because there's a high likelihood that those will be all self-similar: in other words, the within-set diversity of the results would be low.  In the context of search and recommendations, this represents a poor use of the available screen real estate, for if one guess is wrong, all results are useless.  We propose that this finding generalises to interactive text-to-image systems, and furthermore suggest that maximising within-set diversity alongside whatever measure(s) of quality are useful in context should -- in theory -- increase the potential for problem reframing and/or transformationally creative output.  

With those assumptions in mind, it becomes necessary to precisely operationalise within-set diversity for an interactive text-to-image context. The field of quality-diversity algorithms \cite{pugh2016quality}, a form of multi-objective optimisation which has been extensively applied before in computational creativity \cite{zammit2022seeding,mccormack2023creative,demke2023transformational}, would seem to be somewhere to look for inspiration, yet in those cases ``diversity'' is typically measured along several domain-specific and pre-defined behavioural variables: they offer no general measure of diversity that might be applicable in our context. Some generalisable ways of measuring diversity have already been proposed in the literature \cite{naeem2020reliable}, yet most of them either require a ground truth (a.k.a. access to a specific test dataset) for comparison, and/or are very expensive to compute.  In today's era of large pre-trained generative models, access to such a ground truth cannot be assumed, and there is a need for a scalable, general, dataset-blind measure of within-set diversity.

To overcome these issues, we propose and compare two versions of a more relaxed approach to estimating within-set diversity that can be computed quickly and without knowing the distribution of the training data. Our approach is instead based on general pre-trained network mappings. Having no ground truth to evaluate our own measures, we also propose an approach for generating artificial data which we would expect a-priori to exhibit a pattern of relative diversity levels, allowing us to check whether the proposed methods align with our expectations. We argue that our proposed measures are more useful than the SOTA in terms of practicality, particularly in the domain of high-quality interactive image generation in computationally creative contexts.


\section{Methods}


When deep neural networks are trained for image classification or similar tasks, the data from an image flows from each layer to the next  as a tensor of values usually referred to as layer ``activations''. These activations contain information about the different characteristics of each image, and are in turn interpreted by the following layer. Since activations are learnt to be useful for performing the task for which the network was trained -- general purpose image recognition, generation, or segmentation, for example -- then pre-trained networks are often ``cropped'' at certain layers, allowing those layer activations to be used as the input to train (typically smaller) networks for different purposes \cite{kora2022transfer}.

This idea has been exploited for other uses, such assessing the quality of image generators using FID \cite{heusel2017FID}. This method uses the second-to-last layer activation of the general-purpose pre-trained image network InceptionV3 \cite{szegedy2016inceptionV3} as latent variables, effectively casting them as constituting a ``conceptual space'' of all natural images \cite{boden2004creative}. FID then compares a test set of real images to a set of generated ones, with a ``perfect'' score of 0 indicating that the distribution of features in the generated images is identical to those of the ``real'' ones. Specifically, under a normality hypothesis, the Frech\'et Distance between the empirical distributions of the latents can be computed explicitly, giving a good proxy for the quality of the generative process.

While this provides a reliable assessment of the ability of a generator to match a dataset, it has two drawbacks. Firstly, that diversity cannot be measured directly (and in fact moving away from the original latent distribution by becoming ``more diverse'' will produce worse FID scores). And secondly, this method requires a ground-truth distribution for computing (a.k.a. a dataset of all relevant ``real'' images), which as previously discussed is inconvenient for our purpose.

Nonetheless, the idea of analyzing an image dataset through the latent space of a pre-trained network can still be of use. By analyzing the empirical probability distribution of a set of generated images, we may get an idea of diversity by looking at its entropy, which has been widely used as a diversity index \cite{jost2006entropy} in other fields. Where FID computes quality as the distance between the distributions of a generated set and a ground truth in a latent space, we instead seek to assess diversity as the entropy of the generated set's same latent variables. 

We describe two approaches to doing so below, detailing how to tractably approximate entropy in a co-creative case. Both measures are ``truncated'' in that they use approximate measures of entropy in order to avoid the requirement of having at least as many samples (as in generated images) as the dimensionality of the latent space, which isn't feasible in most interactive contexts. The first, Truncated Inception Entropy \cite{ibarrola2022cicada}, is a measure of diversity using the same latent space as in the broadly-adopted FID, on the motivation that if the second-to-last layer of the Inceptionv3 model is a good proxy for image features relevant to quality, it should likely be similar with respect to diversity.  In the second, Truncated CLIP Entropy, we instead explore the use of Contrastive Language-Image Pre-Training (CLIP \cite{radford2021clip}), a multi-modal embedding of both text and images. This is motivated by the assumption that in some use-cases, diversity in a ``semantic'' space that can embed both prompt and images may be more relevant than the features of a general-purpose image model.

\subsection{Truncated Inception Entropy}

Let us consider a function $f$ that maps images into a latent space $\mathcal{Z} \subset \mathbb{R}^D$, in such a way that the points have a normal distribution $\pi_f \sim \mathcal{N}(\mu, \Sigma_i)$ in $\mathcal{Z}$. This normality assumption is the same one used when computing FIDs, where $f$ is a truncated version of the InceptionV3 network on the last layer, resulting in $D=2048$.\footnote{This choice of layer from which to extract activations is the standard for FID, yet other intermediate layers might be considered provided a reliable way to deal with their high dimensionality. Further exploration is required.}

Under this hypothesis, we could assess the diversity of a given set of images by computing the differential entropy $h$ \cite{shannon2001mathematical} of such normal distribution, defined as
\begin{align}
h(\pi_f) = -\mathbb{E}\log(\pi_f) = \frac{1}{2}\log\det(2\pi e \Sigma_i). \label{eqn:entropy}
\end{align}

When the number of samples $N$ in a set of images $A$ is smaller than the dimension $D$ of the latent space, the empirical approximation $\hat{\Sigma}_i$ of $\Sigma_i$ is singular, meaning that the determinant is null and hence the latter computation unfeasible. To overcome this, it has been proposed \cite{ibarrola2022cicada} that a truncated version of entropy can be used, defined as
\begin{align}
\text{TIE}_K(A) \doteq \frac{K}{2}\log(2\pi e) + \frac{1}{2}\sum_{k=1}^{K}\log\lambda^{(i)}_k,\label{eqn:tie}
\end{align}
where TIE denotes Truncated Inception Entropy, and $\{\lambda^{(i)}_k, k=1, \ldots, K\}$ is the set of the $K$ largest eigenvalues of $\hat{\Sigma}$. Note that $K=D$ would make the TIE equivalent to Equation (\ref{eqn:entropy}), but choosing a smaller value for $K$ would let us compare diversities of smaller sets of images.

\subsection{Truncated CLIP Entropy}

The InceptionV3 network was trained as a classifier over the ImageNet database \cite{deng2009imagenet}. Since then, new pretrained networks have been made available, such as CLIP, in which images and text are encoded together in a shared latent space $\mathcal{Z} \subset \mathbb{R}^{512}$. This is done in such a way that text or images with the same semantic characteristics are grouped together, which may be a useful feature of a space in which we want to calculate within-set diversity.

In an analogous way as with the TIE, we may consider a set $A$ of $N$ images and $g(A) \doteq \{g(a), a\in A, g(a)\in\mathbb{R}^{512} \}$ set of latent CLIP representations of the images (where $g$ denotes the CLIP image encoder). From this set, we can calculate the empirical covariance matrix $\hat{\Sigma}_c\in\mathbb{R}^{512\times 512}$, and subsequently its $K$ largest eigenvalues $\{\lambda^{(c)}_k, k=1, \ldots, K\}$ to compute the Truncated CLIP Entropy (TCE) as
\begin{align}
\text{TCE}_K(A) \doteq \frac{K}{2}\log(2\pi e) + \frac{1}{2}\sum_{k=1}^{K}\log\lambda^{(c)}_k.\label{eqn:tce}
\end{align}

Note that while the computation is the same as that of the TIE, the values are not directly comparable, since the spaces in which the InceptionV3 and CLIP latents are defined are different (hence the supra-index notation on the eigenvalues).

\subsection{Open-source Implementation}

The code (Python3) for trying the measures described here is freely available, and may be installed using pip

\smallskip
\noindent
\texttt{\$ pip install image-diversity}
\smallskip

\noindent
and tested by running

\smallskip
\noindent
\texttt{\$ python3 image\_diversity <path/to/dir>}
\medskip

\noindent
where \texttt{<path/to/dir>} is a path to a directory containing a set of images to be evaluated.

More details on installation and usage can be found at \href{https://github.com/fibarrola/image_diversity}{https://github.com/fibarrola/image\_diversity}

\begin{figure*}
    \centering
    \includegraphics[width=\textwidth]{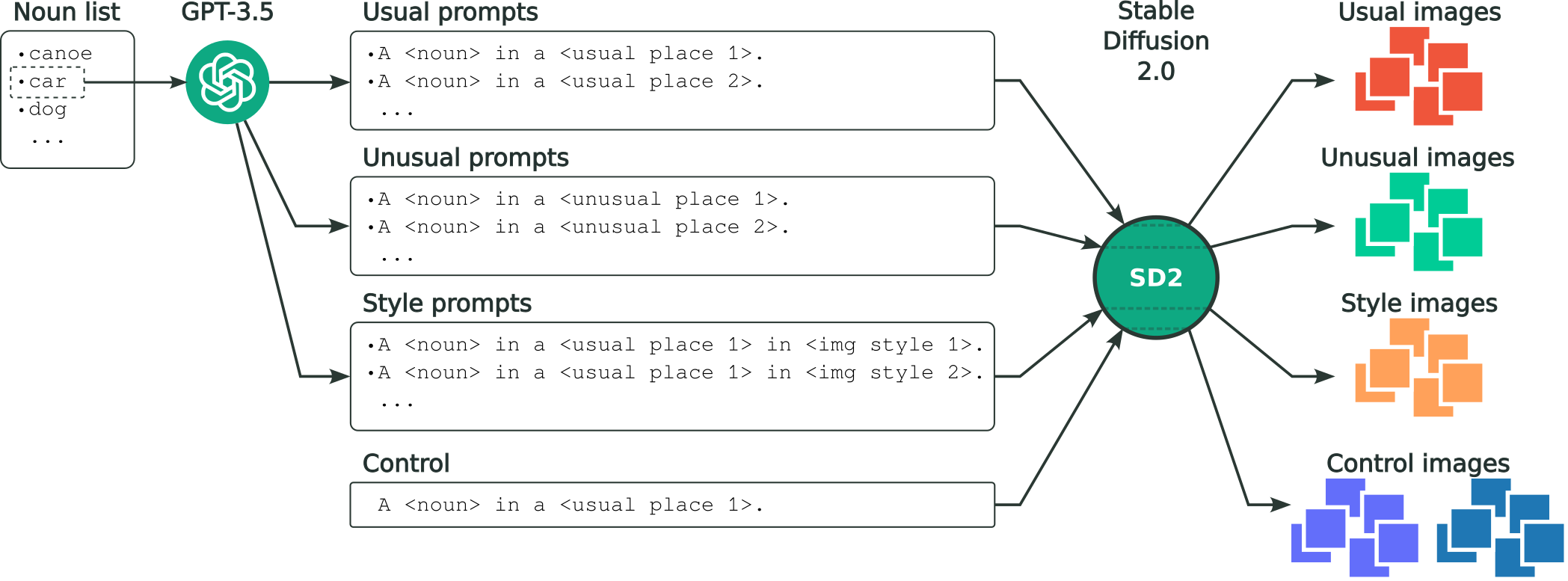}
    \caption{Image set generation process for diversity evaluation.}
    \label{fig:diversity_set_gen}
\end{figure*}


\section{Experiments}


Comparing diversity as estimated by either TIE or TCE is a non-trivial problem, given that there is no ground truth on what the diversity of set of images ``should be''. Or rather, what makes a set of images more or less diverse than another. We are currently in the process of designing a set of human-subjects evaluations to compare different versions of these measures on the degree to which they align with human evaluation. A key challenge in that experimental design is what exactly to ask people to do, rate, or judge in order to validate our diversity measures and our hypothesis that generator diversity facilitates output creativity.  For this paper, however, we present a series of in-silico experiments.  We have built sets of images using different processes that we judge should lead to more or less diversity, and confirmed whether our diversity measures reflect those a-priori assumptions.  This approach is consistent with past experiments on computational diversity measures, such as in the domain of text documents \cite{bache2013text}. Specifically, we automated the generation of sets of prompts that vary in content and style in ways that are both congruous and incongruous.

This was carried out using GPT-3.5 \cite{brown2020gpt3} to generate different text prompts, which were in turn used to generate five datasets: Control with Low Noise (a fixed prompt with small variations in random generative components), Control with High Noise (a fixed prompt with large variations in random generative components), Usual (a given object in different places it might be), Unusual (a given object in places it would not be) and Style (a given object in a Usual place rendered in different visual styles). 

\begin{figure*}
    \centering

    Control (low noise): ``A canoe in a serene lake.''
    \smallskip
    
    \includegraphics[width=0.22\textwidth]{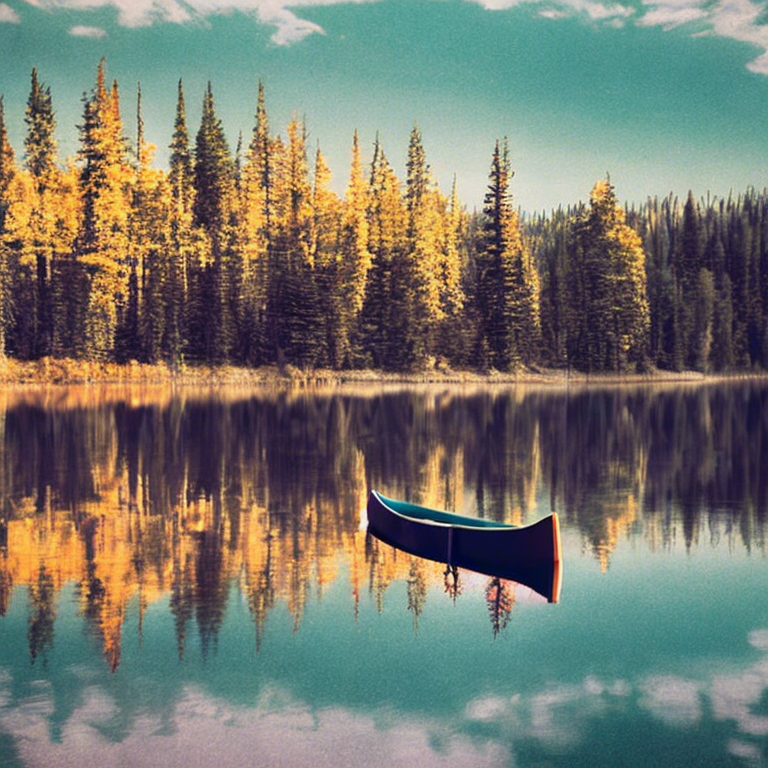}
    \hspace{0.03\textwidth}
    \includegraphics[width=0.22\textwidth]{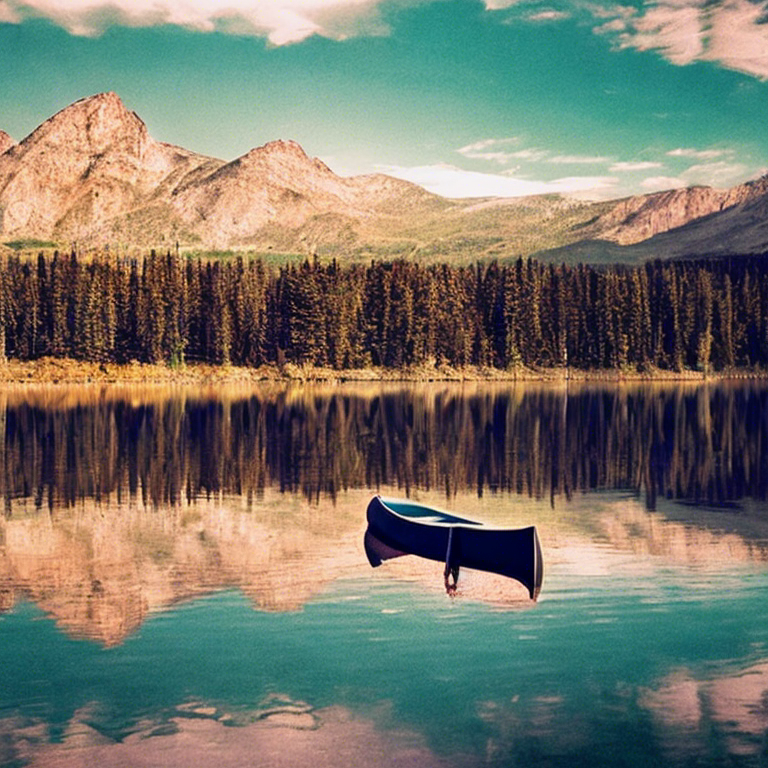}
    \hspace{0.03\textwidth}
    \includegraphics[width=0.22\textwidth]{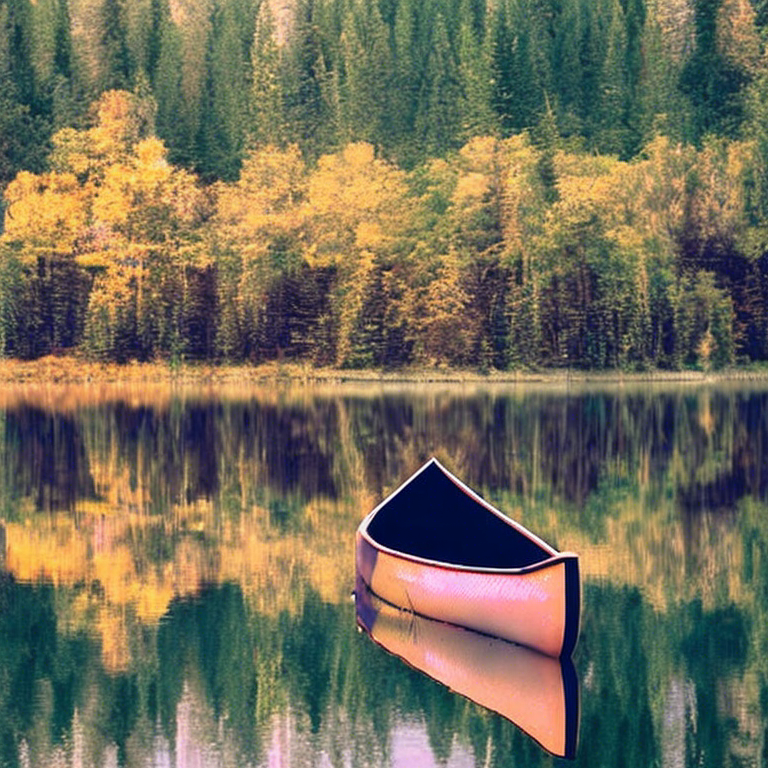}
    \hspace{0.03\textwidth}
    \includegraphics[width=0.22\textwidth]{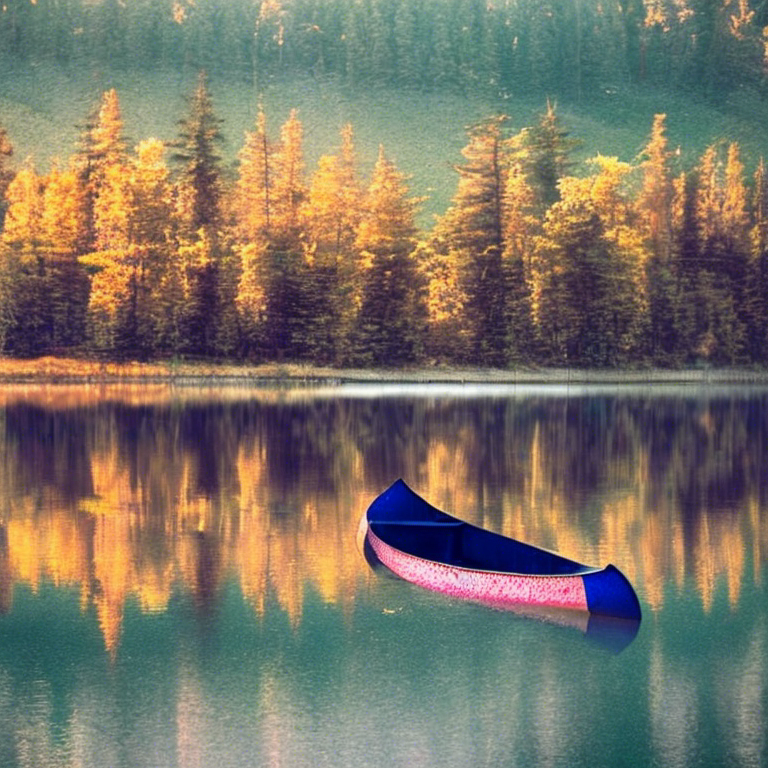}

    \medskip
    Control (high noise): ``A canoe in a serene lake.''
    \smallskip
    
    \includegraphics[width=0.22\textwidth]{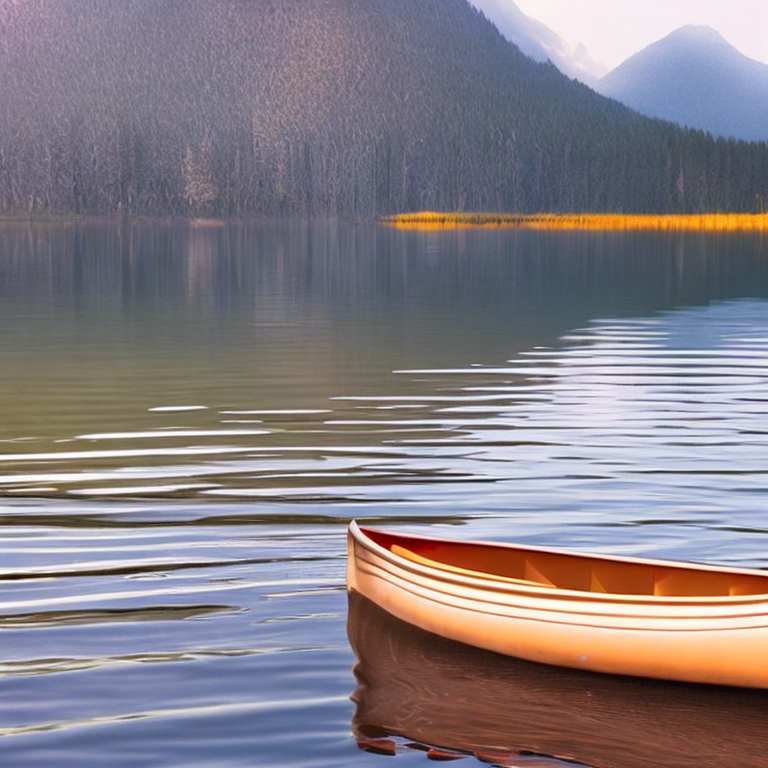}
    \hspace{0.03\textwidth}
    \includegraphics[width=0.22\textwidth]{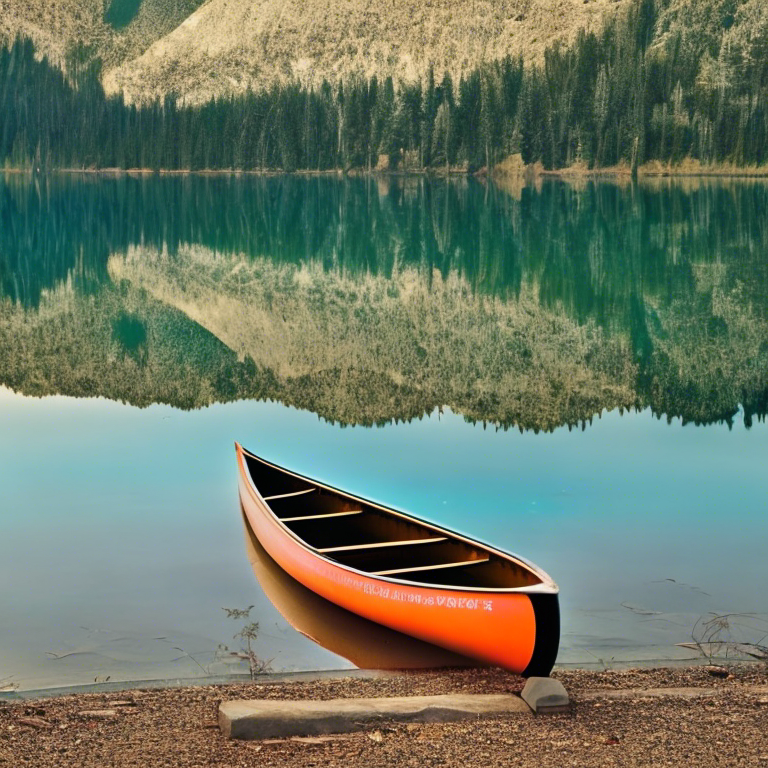}
    \hspace{0.03\textwidth}
    \includegraphics[width=0.22\textwidth]{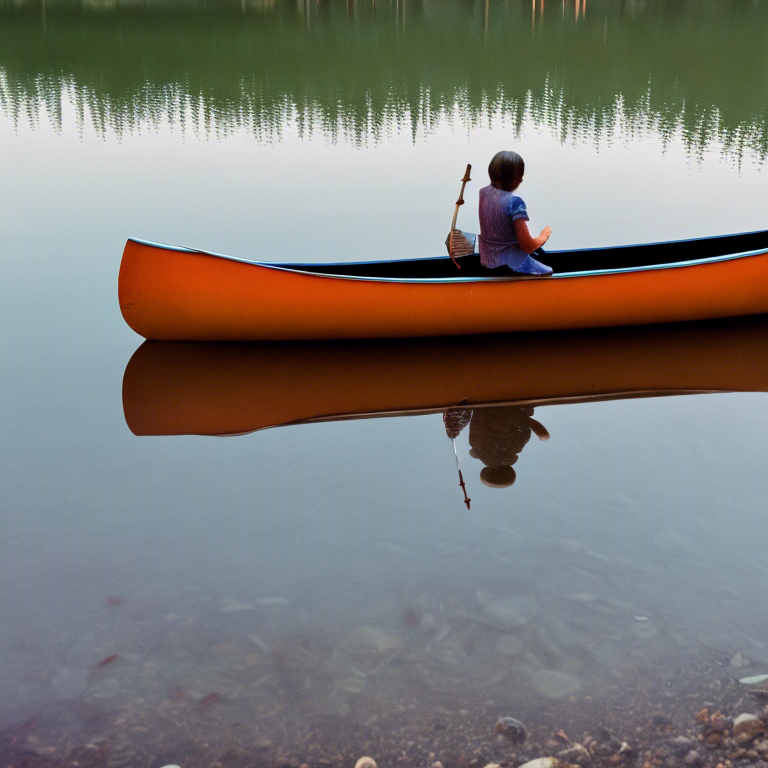}
    \hspace{0.03\textwidth}
    \includegraphics[width=0.22\textwidth]{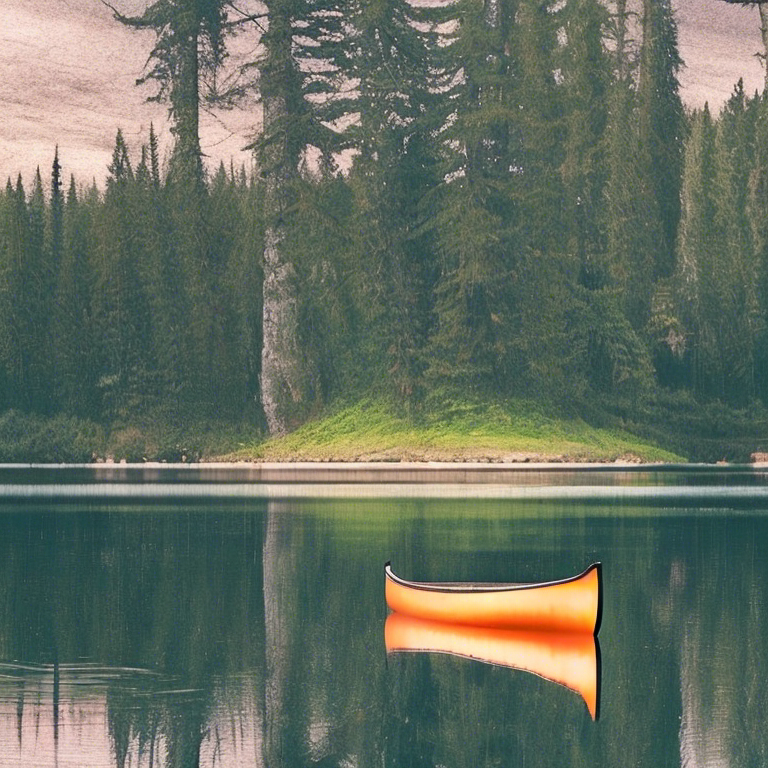}

    \medskip
    Usual: ``A canoe in a $<$usual place$>$''
    \smallskip
    
    \includegraphics[width=0.22\textwidth]{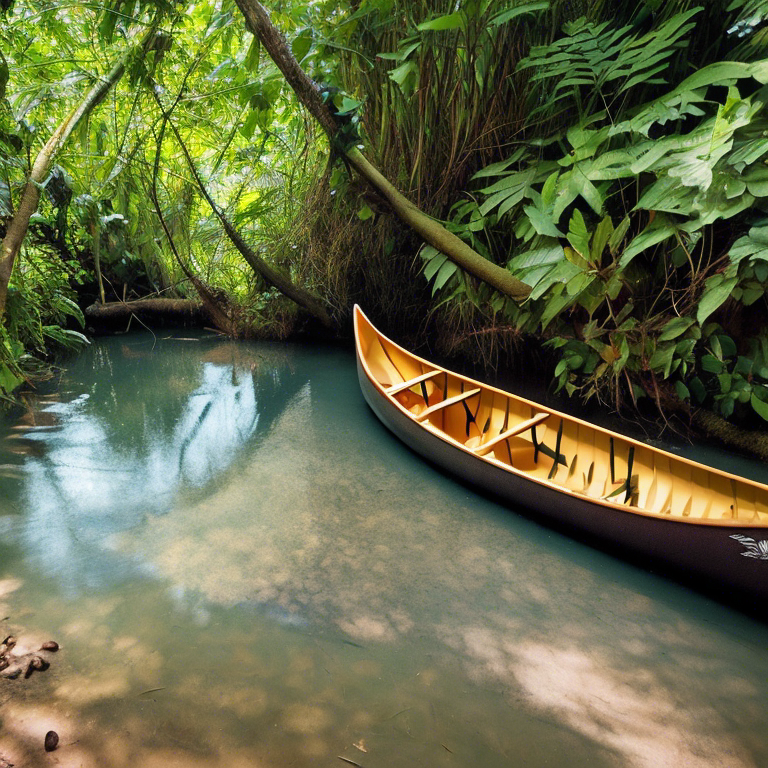}
    \hspace{0.03\textwidth}
    \includegraphics[width=0.22\textwidth]{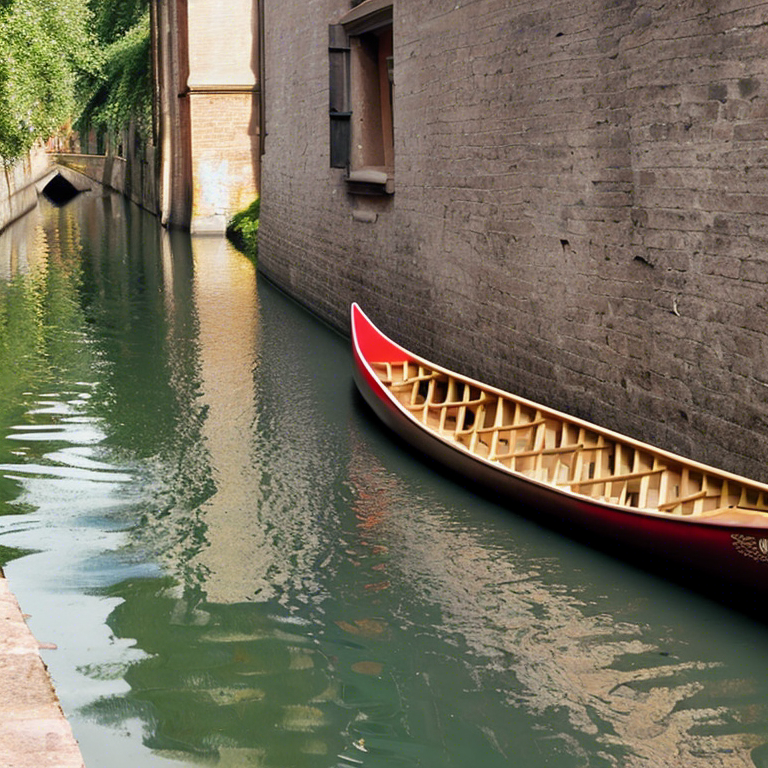}
    \hspace{0.03\textwidth}
    \includegraphics[width=0.22\textwidth]{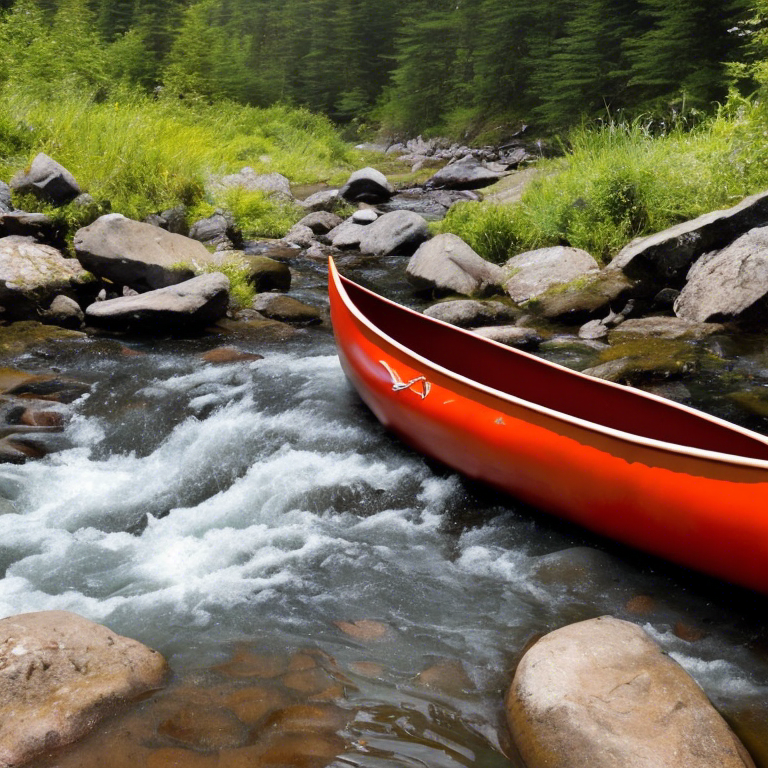}
    \hspace{0.03\textwidth}
    \includegraphics[width=0.22\textwidth]{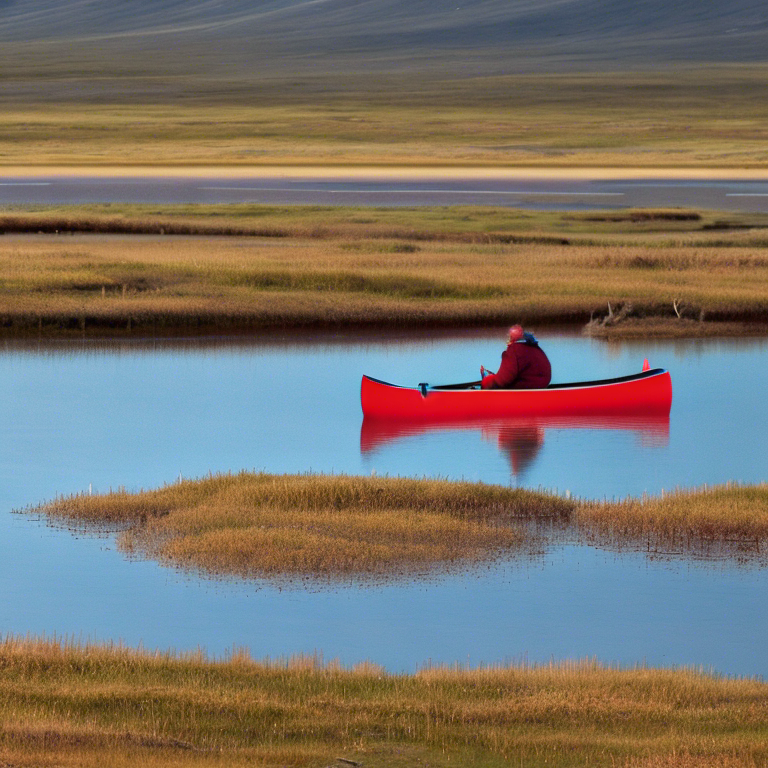}

    \medskip
    Unusual: ``A canoe in a $<$unusual place$>$''
    \smallskip
    
    \includegraphics[width=0.22\textwidth]{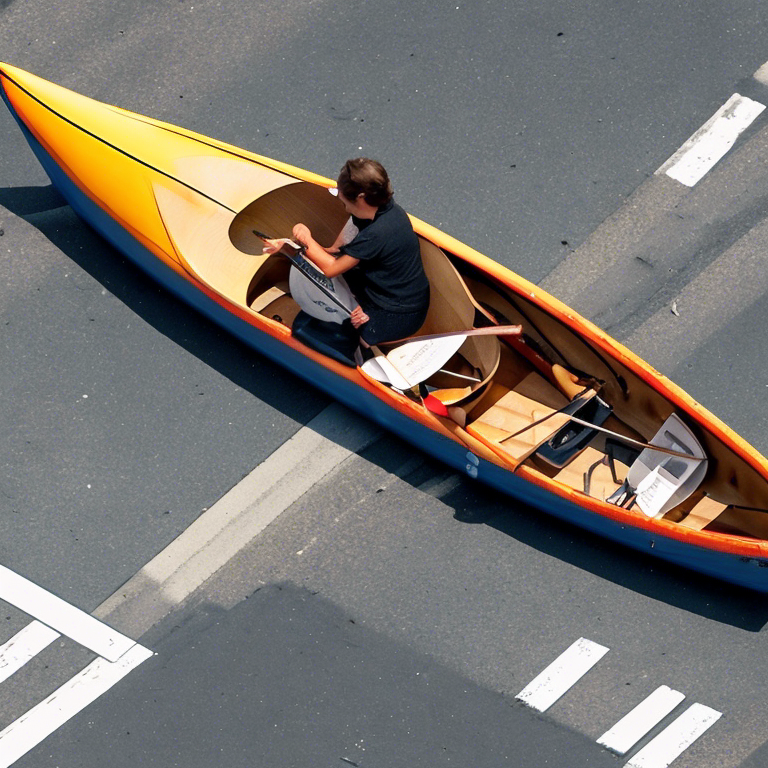}
    \hspace{0.03\textwidth}
    \includegraphics[width=0.22\textwidth]{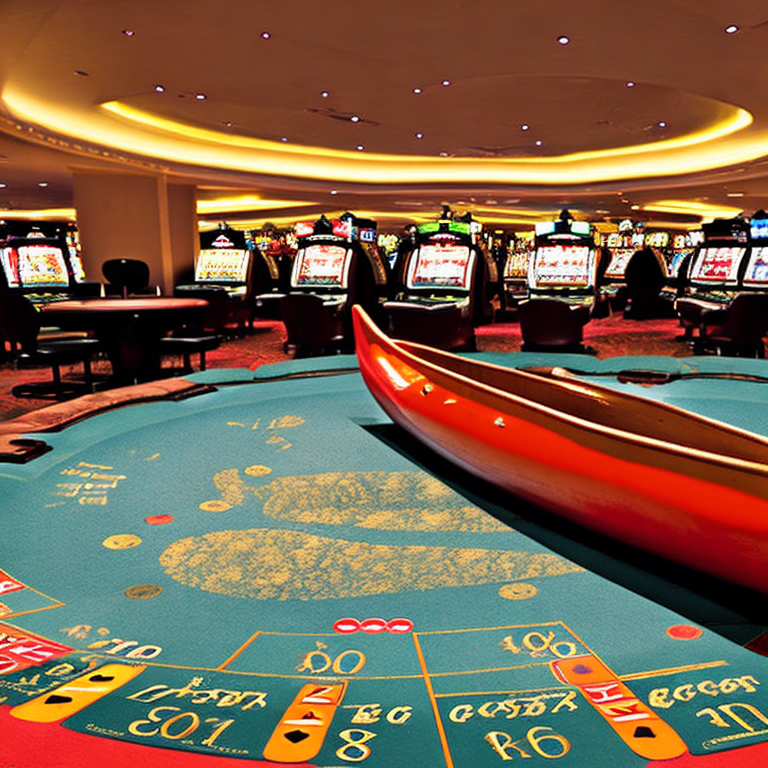}
    \hspace{0.03\textwidth}
    \includegraphics[width=0.22\textwidth]{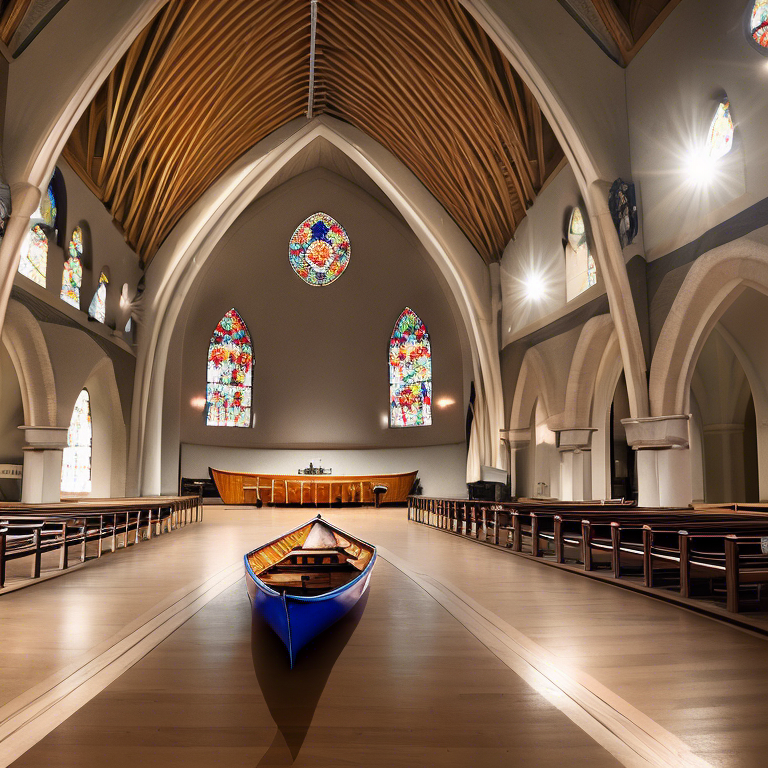}
    \hspace{0.03\textwidth}
    \includegraphics[width=0.22\textwidth]{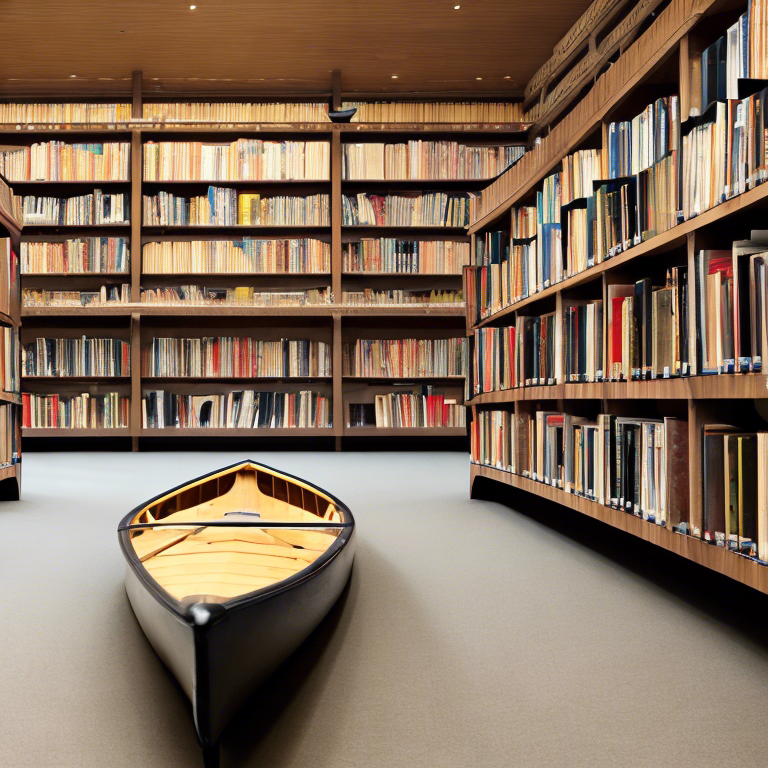}

    \medskip
    Style ``A canoe in a serene lake in $<$image style$>$''
    \smallskip
    
    \includegraphics[width=0.22\textwidth]{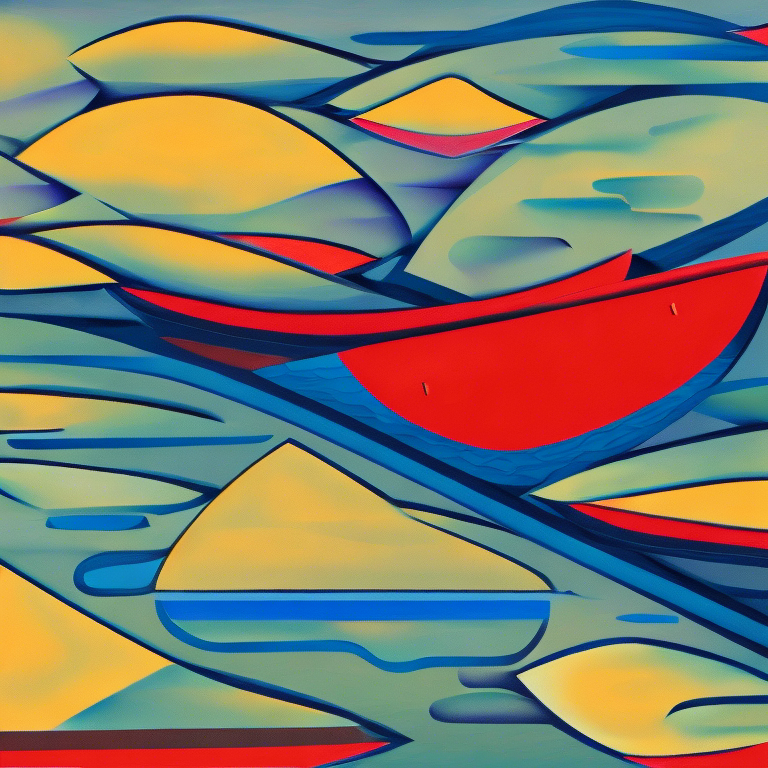}
    \hspace{0.03\textwidth}
    \includegraphics[width=0.22\textwidth]{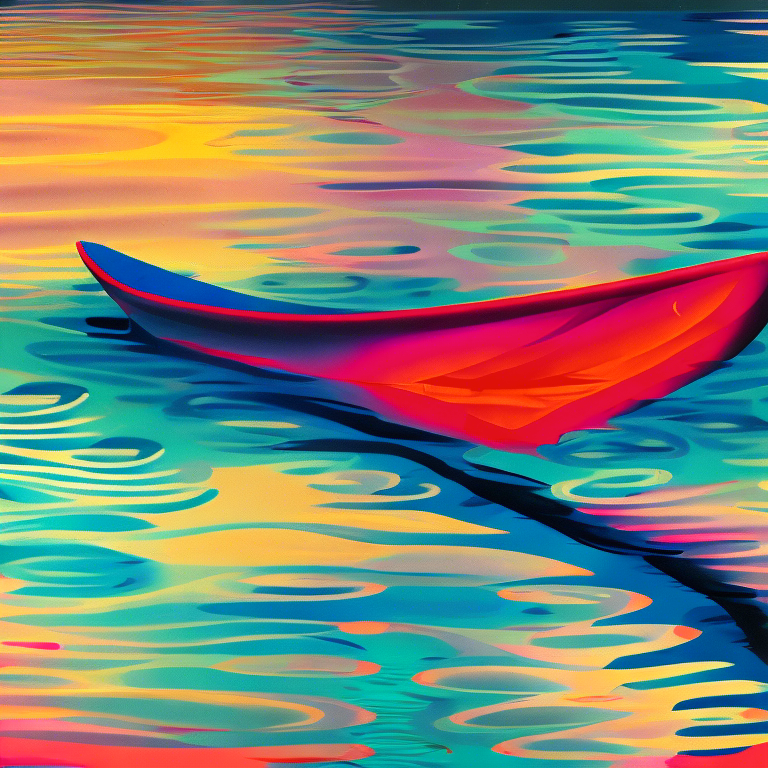}
    \hspace{0.03\textwidth}
    \includegraphics[width=0.22\textwidth]{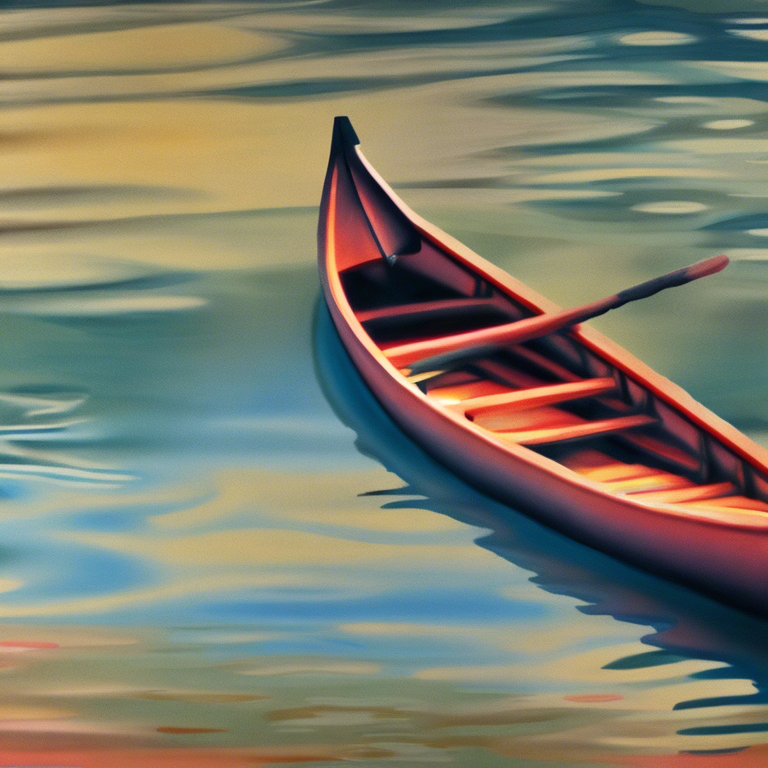}
    \hspace{0.03\textwidth}
    \includegraphics[width=0.22\textwidth]{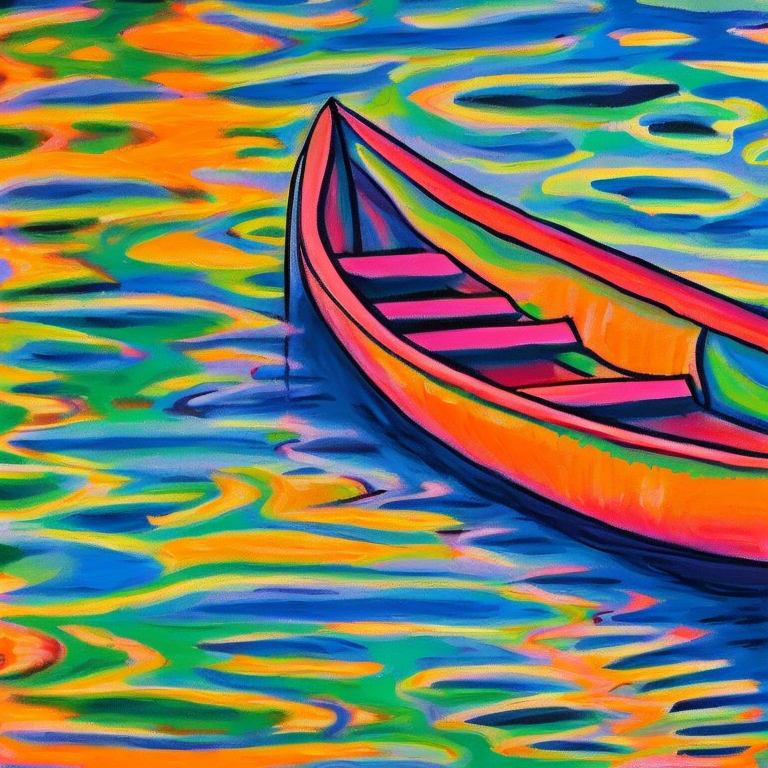}
    
    \caption{Samples of image sets generated with one of three methods to evaluate diversity behaviour.}
    \label{fig:set_samples}
\end{figure*}

If our intuitions about within-set diversity are accurate, two things should occur. Firstly, the low noise Control set should be less diverse than the high noise Control set. Secondly, the low noise Control set should show less diversity than the Usual set, and both of them less than the Unusual set. Finally, the Style set's diversity should be purely visual with low semantic variations, and hence we expect it might be assessed differently by TCE and TIE, due to the latter's presumed greater reliance on visual rather than semantic differences. 

The generative process of the image sets is illustrated in Figure \ref{fig:diversity_set_gen}, and was conducted as follows. We first chose five nouns: \textit{canoe}, \textit{car}, \textit{dog}, \textit{coffee mug} and \textit{pigeon} and then gave the LLM instructions to generate three different sets of prompts as follows:
\begin{description}
    \item[Usual:] Generate a list of 45 places where a [noun] may be. Print as ``A [noun] in $<$place$>$''
    \item[Unusual:] Generate a list of 45 places where finding a [noun] would be absurd. Print as ``A [noun] in $<$place$>$''
    \item[Style:] Generate a list of 45 painting or image styles. Print as ``A [noun] in [place] in $<$style$>$''
\end{description}

In each case [noun] was replaced with one of the five objects in the list above. The \textit{Control} sets did not use an LLM to generate as the prompts were fixed to a single place, chosen to be stereotypical for that object. The prompts in this case were \textit{a canoe in a serene lake}, \textit{a car in a driveway}, \textit{a dog in a backyard}, \textit{a coffee mug in an office}, and \textit{a pigeon in a tree}. These same ``fixed'' places were used for each of the Style prompts. The three varying-prompt sets used the same fixed random parameters to remove that as a source of variance, while the Control sets used 45 instances of varying random components (since with fixed noise all the images would have been identical). In order to further reduce the variance in the low-noise Control set, the parameters were set to vary only 20\%, the effect of which can be observed on the samples in Figure \ref{fig:set_samples}.

Finally, we used Stable Diffusion \cite{rombach2022stablediffusion} to generate sets of 45 image per object, examples of which can be seen in Figure \ref{fig:set_samples}. It can readily be observed that the low-noise Control set show images very similar both visually and in terms of the depicted elements, while the high-noise Control set varies much more visually, yet no more in terms of elements. The Usual set shows some common elements besides the canoe itself, such as water and vegetation, but more variability than the control. In contrast, the Unusual set shows a variety of elements not quite related to each other, from which we should expect a larger diversity. Finally, the Style set is quite consistent in terms of the depicted scene, but is more diverse in terms of geometry and textures.

\begin{figure*}
    \centering
    \includegraphics[width=\textwidth]{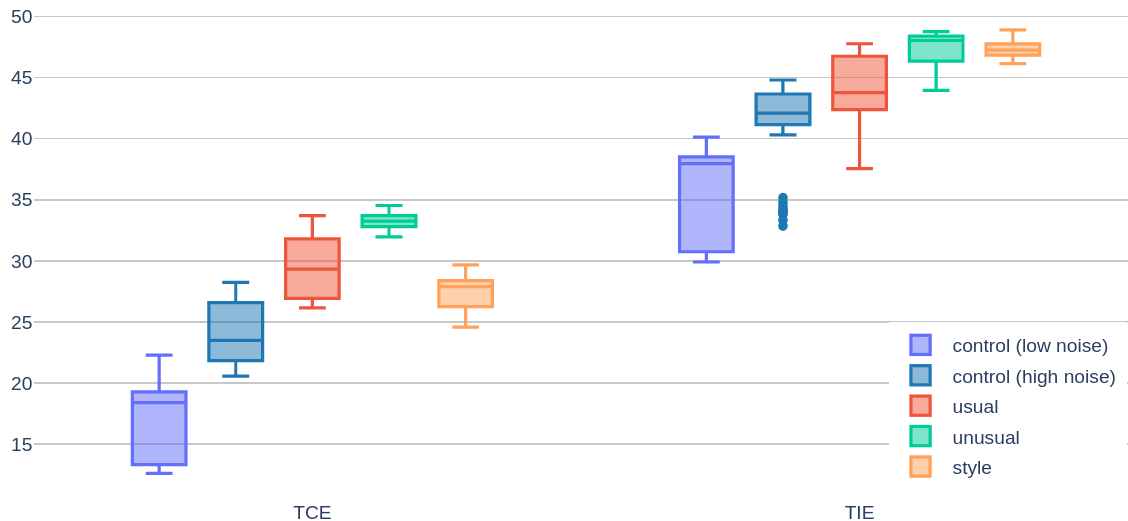}
    \caption{Diversity values (TIE and TCE) using $K=20$ eigenvalues, for sets of images generated with four different criteria. All the between-set differences are statistically significant $(p<0.01)$ except for the TIE for the \emph{unusual} and \emph{style} sets.}
    \label{fig:diversity_boxplot}
\end{figure*}

For each of the 5 objects, we built 10 random subsets of 30 (out of 45) images, and computed the TIE and TCE values, depicted in Figure \ref{fig:diversity_boxplot}. It can be seen that, as expected, the low-noise Control set shows lower diversity than the rest, and the highest diversity scores are observed for the Unusual set with both methods. Unsurprisingly, also, reducing the variance of the input noise (in the Control set) reduces the diversity of the output. However, the TIE marks the Unusual and Style sets as having comparable diversity, significantly greater than the mild variations in the Usual and Noise groups, whereas the TCE tells a different story. In this case, the variations in visual style carry a lower weight than those of the elements composing the image, meaning that TIE and TCE are accounting for diversity in two different senses. This comports with our expectation that TCE ``weights'' semantics higher in its accounting of image diversity.

\subsection{Text diversity}

\begin{figure}
    \centering
    \includegraphics[width=0.5\textwidth]{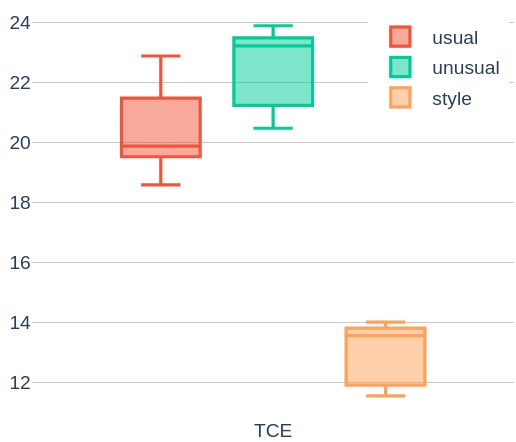}
    \caption{TCE using $K=20$ eigenvalues, for sets of text prompt generated with three different criteria. All the between-set differences are statistically significant $(p<0.01)$.}
    \label{fig:diversity_text}
\end{figure}

As mentioned before, the CLIP network on which TCE is based embeds both images and text in a shared latent space. This means that TCE can be computed (as in \ref{eqn:tce}) on the CLIP latents of a set of prompts directly, without requiring that they be first converted into images. This suggests a potential application of TCE to text diversity, which may be useful by itself or as a comparison to image diversity.

While a rigorous evaluation would be required before claiming that TCE could be used on text to assess semantic diversity in any useful way, we conducted a preliminary experiment of computing the TCE over the prompts (see Figure \ref{fig:diversity_set_gen}) used in our previous experiments, with the exclusion of the Control sets for which the prompts were all identical.

The results are depicted in Figure \ref{fig:diversity_text} and are broadly in line with those obtained for images for the Usual and Unusual groups, with the latter being higher. It can also be observed that there is a very wide gap between these and the Style set, which was not observed in the case of images. This makes sense, as the prompt texts only differed by that one or two style words, making them semantically quite similar, while that one word had a large effect on the visual content of the image, at least according to TIE. This again provides some early evidence to support our diversity measures as capturing a quantity of potential interest to the developers of co-creative systems and other interactive applications of generative AI.


\section{Discussion and Conclusions}


We proposed a method to assess diversity in image datasets that is agnostic to training data and simple to compute. The method was compared to its analogous using another network's latent space, and results show both variants to align well with expected outcomes. Furthermore, it has been shown that the different networks assess diversity in different senses, meaning that they might serve for different creative contexts.

Our measures are based on approximations of entropy, and entropic measures of diversity have faced some criticism in other fields, such as in biology \cite{jost2006entropy}. The criticism is that the actual quantity of interest in diversity is how many meaningfully active categories (species in biodiversity, ``features'' in an image) in a sample, not the amount of information required to identify which category a randomly-selected sample belongs to. Qualities such as \textit{balance}, \textit{variety} and \textit{disparity} have been proposed as necessary components of this kind of categorical measure of diversity \cite{stirling2007general}. This approach has been applied to evaluating document diversity using topic modelling to generate the categorical representation \cite{bache2013text}. In the case of image generation, this might suggest an alternative formulation in terms of the number of features identified by some appropriately categorical representation.

While our results are promising, further experiments are needed to fully assess the proposed methods' compliance with expectations in creative computing applications. Particularly, future work shall deal with the validation of these metrics in comparison with human perception, and exploring the use of latent spaces of other pre-trained neural networks. In fact, the possibility of using average pooling for computing FID using intermediate InceptionV3 layers has been proposed, although not properly tested \cite{Seitzer2020FID}, and its usage for computing TIE is thus equally plausible. Using earlier layers in the image encoding network as the latent space in which diversity is calculated could yield a more texturally- or visually- biased measure, which may be useful for some scenarios, although only if some technique like average pooling can be applied to reduce their dimensionality.

Finally, as shown by the preliminary experiments, it is worth noting that TCE might also be used to assess text diversity on account of the CLIP latent space being the same for either text or images. More experiments are needed to properly test whether or not this works reliably in practice, contrasting it with other text diversity assessment methods. Our current research is exploring both the design of those experiments as well as the design of future generative systems aimed at producing small sets of diverse-yet-high-quality responses for use in co-creative systems.


\section*{Acknowledgments}


We would like to acknowledge the Australian Research Council for funding this research (ID \#DP200101059).

\bibliographystyle{unsrt}

\begin{thebibliography}{10}

\bibitem{heusel2017FID}
Martin Heusel, Hubert Ramsauer, Thomas Unterthiner, Bernhard Nessler, and Sepp Hochreiter.
\newblock Gans trained by a two time-scale update rule converge to a local nash equilibrium.
\newblock {\em Advances in neural information processing systems}, 30, 2017.

\bibitem{smith2010analyzing}
Gillian Smith and Jim Whitehead.
\newblock Analyzing the expressive range of a level generator.
\newblock In {\em Proceedings of the 2010 workshop on procedural content generation in games}, pages 1--7, 2010.

\bibitem{preuss2014searching}
Mike Preuss, Antonios Liapis, and Julian Togelius.
\newblock Searching for good and diverse game levels.
\newblock In {\em 2014 IEEE Conference on Computational Intelligence and Games}, pages 1--8. IEEE, 2014.

\bibitem{ibarrola2023affect}
Francisco Ibarrola, Rohan Lulham, and Kazjon Grace.
\newblock Affect-conditioned image generation.
\newblock {\em arXiv preprint arXiv:2302.09742}, 2023.

\bibitem{schon1992designing}
Donald~A Sch{\"o}n.
\newblock Designing as reflective conversation with the materials of a design situation.
\newblock {\em Knowledge-based systems}, 5(1):3--14, 1992.

\bibitem{dorst2015frame}
Kees Dorst.
\newblock {\em Frame innovation: Create new thinking by design}.
\newblock MIT press, 2015.

\bibitem{grace2016surprise}
Kazjon Grace and Mary~Lou Maher.
\newblock Surprise-triggered reformulation of design goals.
\newblock In {\em Proceedings of the AAAI Conference on Artificial Intelligence}, volume~30, 2016.

\bibitem{oppenlaender2022creativity}
Jonas Oppenlaender.
\newblock The creativity of text-to-image generation.
\newblock In {\em Proceedings of the 25th International Academic Mindtrek Conference}, pages 192--202, 2022.

\bibitem{candillier2011diversity}
Laurent Candillier, Max Chevalier, Damien Dudognon, and Josiane Mothe.
\newblock Diversity in recommender systems.
\newblock In {\em Proceedings: The Fourth International Conference on Advances in Human-oriented and Personalized Mechanisms, Technologies, and Services. CENTRIC}, pages 23--29, 2011.

\bibitem{kunaver2017diversity}
Matev{\v{z}} Kunaver and Toma{\v{z}} Po{\v{z}}rl.
\newblock Diversity in recommender systems--a survey.
\newblock {\em Knowledge-based systems}, 123:154--162, 2017.

\bibitem{pugh2016quality}
Justin~K Pugh, Lisa~B Soros, and Kenneth~O Stanley.
\newblock Quality diversity: A new frontier for evolutionary computation.
\newblock {\em Frontiers in Robotics and AI}, 3:40, 2016.

\bibitem{zammit2022seeding}
Marvin Zammit, Antonios Liapis, and Georgios~N Yannakakis.
\newblock Seeding diversity into ai art.
\newblock {\em arXiv preprint arXiv:2205.00804}, 2022.

\bibitem{mccormack2023creative}
Jon Mccormack, Camilo Cruz~Gambardella, and Stephen Krol.
\newblock Creative discovery using quality-diversity search.
\newblock In {\em Proceedings of the Companion Conference on Genetic and Evolutionary Computation}, pages 747--750, 2023.

\bibitem{demke2023transformational}
Jonathan Demke, Kazjon Grace, Francisco Ibarrola, and Dan Ventura.
\newblock Transformational creativity through the lens of quality-diversity.
\newblock In {\em Proceedings of ICCC'23}, 2023.

\bibitem{naeem2020reliable}
Muhammad~Ferjad Naeem, Seong~Joon Oh, Youngjung Uh, Yunjey Choi, and Jaejun Yoo.
\newblock Reliable fidelity and diversity metrics for generative models.
\newblock In {\em International Conference on Machine Learning}, pages 7176--7185. PMLR, 2020.

\bibitem{kora2022transfer}
Padmavathi Kora, Chui~Ping Ooi, Oliver Faust, U~Raghavendra, Anjan Gudigar, Wai~Yee Chan, K~Meenakshi, K~Swaraja, Pawel Plawiak, and U~Rajendra Acharya.
\newblock Transfer learning techniques for medical image analysis: A review.
\newblock {\em Biocybernetics and Biomedical Engineering}, 42(1):79--107, 2022.

\bibitem{szegedy2016inceptionV3}
Christian Szegedy, Vincent Vanhoucke, Sergey Ioffe, Jon Shlens, and Zbigniew Wojna.
\newblock Rethinking the inception architecture for computer vision.
\newblock In {\em Proceedings of the IEEE conference on computer vision and pattern recognition}, pages 2818--2826, 2016.

\bibitem{boden2004creative}
Margaret~A Boden.
\newblock {\em The creative mind: Myths and mechanisms}.
\newblock Psychology Press, 2004.

\bibitem{jost2006entropy}
Lou Jost.
\newblock Entropy and diversity.
\newblock {\em Oikos}, 113(2):363--375, 2006.

\bibitem{ibarrola2022cicada}
Francisco Ibarrola, Tomas Lawton, and Kazjon Grace.
\newblock A collaborative, interactive and context-aware drawing agent for co-creative design.
\newblock {\em arXiv preprint arXiv:2209.12588}, 2022.

\bibitem{radford2021clip}
Alec Radford, Jong~Wook Kim, Chris Hallacy, Aditya Ramesh, Gabriel Goh, Sandhini Agarwal, Girish Sastry, Amanda Askell, Pamela Mishkin, Jack Clark, et~al.
\newblock Learning transferable visual models from natural language supervision.
\newblock In {\em International Conference on Machine Learning}, pages 8748--8763. PMLR, 2021.

\bibitem{shannon2001mathematical}
Claude~Elwood Shannon.
\newblock A mathematical theory of communication.
\newblock {\em ACM SIGMOBILE mobile computing and communications review}, 5(1):3--55, 2001.

\bibitem{deng2009imagenet}
Jia Deng, Wei Dong, Richard Socher, Li-Jia Li, Kai Li, and Li~Fei-Fei.
\newblock Imagenet: A large-scale hierarchical image database.
\newblock In {\em 2009 IEEE conference on computer vision and pattern recognition}, pages 248--255. Ieee, 2009.

\bibitem{bache2013text}
Kevin Bache, David Newman, and Padhraic Smyth.
\newblock Text-based measures of document diversity.
\newblock In {\em Proceedings of the 19th ACM SIGKDD international conference on Knowledge discovery and data mining}, pages 23--31, 2013.

\bibitem{brown2020gpt3}
Tom Brown, Benjamin Mann, Nick Ryder, Melanie Subbiah, Jared~D Kaplan, Prafulla Dhariwal, Arvind Neelakantan, Pranav Shyam, Girish Sastry, Amanda Askell, et~al.
\newblock Language models are few-shot learners.
\newblock {\em Advances in neural information processing systems}, 33:1877--1901, 2020.

\bibitem{rombach2022stablediffusion}
Robin Rombach, Andreas Blattmann, Dominik Lorenz, Patrick Esser, and Bj{\"o}rn Ommer.
\newblock High-resolution image synthesis with latent diffusion models.
\newblock In {\em Proceedings of the IEEE/CVF Conference on Computer Vision and Pattern Recognition}, pages 10684--10695, 2022.

\bibitem{stirling2007general}
Andy Stirling.
\newblock A general framework for analysing diversity in science, technology and society.
\newblock {\em Journal of the Royal Society interface}, 4(15):707--719, 2007.

\bibitem{Seitzer2020FID}
Maximilian Seitzer.
\newblock {pytorch-fid: FID Score for PyTorch}.
\newblock \url{https://github.com/mseitzer/pytorch-fid}, August 2020.
\newblock Version 0.3.0.

\end{thebibliography}

\end{document}